\documentclass[11pt]{article}

\usepackage[preprint]{acl}

\usepackage{times}
\usepackage{latexsym}

\usepackage[T1]{fontenc}

\usepackage[utf8]{inputenc}

\usepackage{microtype}

\usepackage{inconsolata}

\usepackage{graphicx}
\usepackage{xcolor}
\usepackage{enumitem}
\usepackage{multirow}
\usepackage{booktabs}
\usepackage{array}
\usepackage{mdframed}
\usepackage{amssymb}   
\usepackage{amsmath}
\usepackage{bm}
\usepackage{bbm}
\usepackage{subcaption}

\title{How Hard Does It Think? Analyzing Step-Aware Reasoning Energy\\ in LLM Chain-of-Thought Trajectories}

\author{Hui Wei $^\dagger$, Junda Wu $^\ddagger$, Sheldon Yu $^\ddagger$, Sizhe Zhou $^\S$, Yizhu Jiao $^\S$, 
Ming Zhong $^\S$, \\ \textbf{Bowen Jin $^\S$, Tong Yu $^\diamondsuit$, Shijia Pan $^\ddagger$, Jiawei Han $^\S$, Julian McAuley $^\ddagger$}\\[0.8em]
$^\dagger$UC Merced \, 
$^\ddagger$UC San Diego\,
$^\S$UIUC \,
$^\diamondsuit$Adobe Research\\[0.3em]
\texttt{huiwei2@ucmerced.edu}
}

\begin{document}
\maketitle
\begin{abstract}
Understanding how computational effort is allocated across individual chain-of-thought (CoT) reasoning steps remains an open challenge: existing interpretability methods rely on output-level signals or collapse processing depth into a single trajectory-level scalar, leaving step-wise effort opaque.
We propose \textbf{Step-Aware Reasoning Energy (SARE)}, a geometric framework that quantifies effort at the granularity of individual CoT steps via Centered Kernel Alignment (CKA) between Gram matrices of token hidden states across adjacent transformer layers, capturing inter-token relational structure without requiring eigenvector alignment or cluster correspondence. SARE further contextualizes this energy within reasoning's semantic progression by modeling CoT trajectories as transitions among latent semantic states.
Across six reasoning benchmarks and three open-weight LLMs, we find that reasoning energy is highly non-uniform across step types, exhibiting phase-like transitions invisible to trajectory-level metrics; incorrect trajectories show systematically lower energy at critical reasoning junctions; and SARE-based features match or outperform output-based confidence baselines in most settings, indicating that internal geometric dynamics encode predictive information beyond surface-level signals.
\end{abstract}
\section{Introduction}

Chain-of-Thought (CoT) prompting \citep{wei2022chain} has become one of the most effective techniques for eliciting multi-step reasoning from large language models(LLMs), substantially improving performance on mathematical, logical, andcommonsense benchmarks. By generating explicit intermediate reasoning steps, CoT provides a window into the model's reasoning process. Yet despite this apparent transparency, we have little understanding of how computational effort is actually allocated across those steps (e.g., which steps demand deep internal processing, and which are resolved trivially), leaving the black-box nature of LLM reasoning largely intact.

Existing attempts to explain CoT reasoning operate primarily at the surface level. Output-based approaches examine token log-probabilities \citep{hwang2026oops} or train classifiers on trajectory text \citep{madaan2023self}, but treat the transformer's internal computations as opaque. Interpretability research has begun
probing transformer internals \citep{belrose2023eliciting, chuang2023dola}, yet typically at the level of individual tokens or collapsed aggregate representations, which is too coarse to capture the computational dynamics of entire reasoning steps.
Recently, \citet{chen2026think} propose the Deep Thinking Ratio (DTR) to measure reasoning effort via token-level layer-wise prediction stability, but DTR aggregates depth into a single trajectory-level scalar and treats tokens independently, discarding the relational structure among tokens within a step. Neither line of work adequately connects the model's layer-wise computation to the semantic progression of the reasoning chain.

To bridge this gap, we propose \textbf{Step-Aware Reasoning Energy (SARE)}, a
geometric framework that quantifies computational effort at the granularity of
individual CoT reasoning steps. For each step and each pair of adjacent transformer
layers, we compute Centered Kernel Alignment (CKA) \citep{kornblith2019similarity}
between Gram matrices constructed from token hidden states, measuring how much the
step's internal token relationship geometry reorganizes during the forward pass. A
step that continues to reorganize across many layers reflects genuine computational
effort; one that stabilizes early indicates minimal processing. Unlike token-level
depth measures, CKA operates on pairwise token similarity structure, preserving
inter-token relational information without requiring eigenvector alignment or
cluster correspondence across layers. We further contextualize SARE within the
semantic trajectory of reasoning by modeling CoT steps as transitions among latent
semantic states identified via unsupervised clustering, enabling joint analysis of
how energy varies across semantic roles and how it evolves across the chain.

We evaluate SARE across six reasoning benchmarks spanning mathematical, commonsense,
and multi-hop domains (i.e., GSM8K \citep{cobbe2021training}, MATH \citep{hendrycks2021measuring}, CSQA \citep{talmor2019commonsenseqa}, StrategyQA \citep{geva2021did}, HotpotQA \citep{yang2018hotpotqa}, and MuSiQue \citep{trivedi2022musique}) using three open-weight LLMs: LLaMA-3.2-3B \citep{grattafiori2024llama}, Phi-4-mini \citep{abouelenin2025phi}, and Gemma-3-4B \citep{gemmateam2025gemma3technicalreport}. Our analysis addresses two core research questions: 

\begin{itemize}
    \item \textbf{RQ1:} Do different semantic reasoning states exhibit distinct
    step-level reasoning energy profiles, and do these profiles differ between
    correct and incorrect trajectories?
    \item \textbf{RQ2:} Can step-level energy dynamics predict reasoning failures
    without access to ground-truth labels?
\end{itemize}

Our results reveal three consistent findings. First, reasoning energy is highly non-uniform across semantic step types: early setup and final synthesis steps anchor the energy extremes while mid-trajectory factual retrieval steps exhibit stable, moderate energy, exposing a structured phase-like allocation of computational effort invisible to trajectory-level metrics. Second, incorrect trajectories are associated with lower reasoning energy at specific reasoning junctions, particularly in final verification and compositional reasoning states. Third, SARE, combined with token count as a complementary signal, match or outperform output-based confidence baselines including token log-probability, entropy, and perplexity on most evaluated benchmarks and models.

In summary, our primary contributions are:
\begin{itemize}
    \item We propose SARE, a geometry-grounded framework that quantifies step-level reasoning effort in CoT trajectories via CKA on token hidden state Gram matrices, capturing inter-token relational dynamics that token-level measures discard.
    \item We establish that reasoning energy is structured and non-uniform across semantic step types, and that incorrect trajectories are consistently associated with lower energy at critical reasoning junctions.
    \item We show that SARE-based features are competitive with or superior to output-based confidence baselines for offline reasoning failure detection across diverse benchmarks, models, and reasoning domains.
\end{itemize}
\section{Related Work}

\subsection{Understanding and Explaining Chain-of-Thought Reasoning}
The emergence of CoT prompting \citep{wei2022chain} and its variants has sparked extensive research into how LLMs solve complex multi-step problems. While empirical results show that CoT substantially enhances reasoning performance, understanding the internal mechanism behind this capability is an active area of study. A significant portion of existing work focuses entirely on the final representations or the generated output layer. For instance, methods like Self-Consistency \citep{wang2022self} sample multiple reasoning paths and aggregate semantic clusters over the final answers to increase reliability. Parallel efforts have analyzed reasoning errors by examining token log-probabilities \citep{kauf2024log} or by classifying reasoning types directly from the textual output trajectory \citep{madaan2023self}. Although these approaches offer practical ways to evaluate the consistency of a reasoning path, they operate entirely on the surface level, leaving the step-to-step computational effort within the deep layer representations of the model unexplored. As a result, surface-level methods fall short in pinpointing the exact internal location where an error begins.

\subsection{Internal Representation Analysis in Transformers}
Analyses of transformer internal mechanics have traditionally investigated how structural information is built progressively across layers. Tools such as ``tuned lenses'' \citep{belrose2023eliciting} and early exiting strategies \citep{schwartz2020right, teerapittayanon2016branchynet} attempt to map hidden states to final vocabulary predictions or determine when computation can be terminated. Recent studies, such as DoLa \citep{chuang2023dola}, focus on decoding strategies that optimize information difference between specific intermediate layers to mitigate hallucinations. However, these methods primarily focus on trajectory-level representation or trace representations per-token across layers. A token-level lens often fails to capture the aggregate semantic value of a contiguous reasoning step (e.g., an equation or a logical deduction). 

Our work bridges the gap between text-level CoT analysis and token-level layer probing. By defining \textbf{Step-Aware Reasoning Energy (SARE)}, we aggregate representational transitions across the specific span of tokens corresponding to a reasoning step. Furthermore, we cluster the final representations into distinct \emph{reasoning states}, moving beyond raw token probabilities. This novel approach quantifies the computational effort of step transitions (measured via layer-wise alignment) and grounds it in the broader semantic topology of CoT reasoning (measured via state transitions based on the final layer). Consequently, our framework provides multiple dimensions: stepwise representation changes and semantic state evolution, to explain and evaluate CoT reasoning capabilities.

\section{Preliminaries: Modeling Reasoning as State Transitions}

We first formalize the CoT trajectory as a stochastic process to analyze the evolution of internal reasoning logic.

\subsection{Step-Aware Formalization}
A CoT reasoning trajectory $\mathcal{T}$ is segmented into a sequence of $T$ discrete textual reasoning steps: $\mathcal{T} = [s_1, \ldots, s_T]$, where each step $s_t$ ($t = 1, \dots, T$) consists of $n_t$ tokens. To capture the internal semantic relationships among tokens at a given depth, we define the Gram matrix $\bm{G}_t^{(l)} = \bm{H}_t^{(l)}(\bm{H}_t^{(l)})^\top$, where $\bm{H}_t^{(l)} \in \mathbb{R}^{n_t \times d}$ denotes the matrix of token hidden states for step $s_t$ at layer $l$.

\subsection{Semantic State Clustering} \label{sec:clustering}
Following \citet{yu2025explainable}, we interpret the progression of reasoning steps as transitions among latent semantic states. Each step $s_t$ is represented by a spectral embedding derived from the eigenvalue spectrum of its cumulative token Gram matrix, computed from the last-layer hidden states of the LLM. These embeddings are then grouped via $K$-Means clustering  to infer reasoning clusters (macro-states) $C \in \{C_1, \dots, C_K\}$, each capturing a distinct conceptual function such as problem framing, intermediate verification, or factual retrieval. Each step is assigned a hard cluster label, and the resulting cluster sequence defines the trajectory's state sequence for downstream analysis. Full implementation details are provided in Appendix~\ref{sec:clustering_details}.

\subsection{Markovian Transition Framework}
We model the sequence of reasoning clusters as a first-order Markov chain. The dynamics of the reasoning process are governed by a transition probability matrix $P$, where the probability of transitioning from cluster $C_i$ to $C_j$ is defined as:
\begin{equation}
    P_{ij} = P(s_{t+1} = C_j \mid s_t = C_i)
\end{equation}
This formulation allows us to track cross-step energy velocities $\Delta E(s_{t} \to s_{t+1})$ and analyze the stability of the reasoning trajectory through the lens of state-space transitions. In practice, these cross-step energy dynamics are operationalized in our downstream analysis through the \emph{volatility}, \emph{peaks}, and \emph{valleys} statistics in the trajectory-level feature vector (Section~\ref{sec:experiment_design}), which collectively capture the magnitude and direction of energy changes between consecutive steps.

\section{Quantifying Reasoning Energy via Geometric Dissimilarity} \label{sec:CKA}
Having established the semantic structure of reasoning trajectories, we now turn to measuring the computational effort expended within each step.

\subsection{Layer-to-Layer Dissimilarity via Centered Kernel Alignment (CKA)}

A key insight from mechanistic interpretability is that transformer layers do not process all tokens equally: some tokens require sustained representational revision across many layers before their contextual role is resolved, while others stabilize early and require little further computation \citep{chuang2023dola, chen2026think}. We operationalize this observation at the \textit{step level}: a reasoning step that undergoes substantial reorganization of its internal token relationship geometry across consecutive layers is one the model actively ``works on,'' reflecting genuine computational effort. Conversely, a step whose geometry stabilizes rapidly indicates that the model resolved it with minimal processing. Crucially, this geometric perspective operates on raw hidden states rather than output-level signals, capturing computational effort that may never surface in the model's token predictions.

\paragraph{Formulation.}
Specifically, We apply CKA \citep{kornblith2019similarity} to the centered Gram matrix \(\tilde{\bm{G}}_t^{(l)} = \bm{M}_{n_t} \bm{G}_t^{(l)} \bm{M}_{n_t}\), where \(\bm{M}_{n_t} = \bm{I}_{n_t} - \frac{1}{n_t}\mathbf{1}\mathbf{1}^\top\) is the standard centering matrix and \(n_t\) is the number of tokens in step \(s_t\). $\bm{I}_{n_t} \in \mathbb{R}^{n_t \times n_t}$ denotes the
identity matrix and $\mathbf{1} \in \mathbb{R}^{n_t}$ is an all-ones vector.
The layer-wise dissimilarity score for step $s_t$ between adjacent layers $l$ and $l+1$ is then:
\begin{equation}
    D_t^{(l)} = 1 - \mathrm{CKA}\big(\tilde{\bm{G}}_t^{(l)},\, \tilde{\bm{G}}_t^{(l+1)}\big),  \text{for } l = 1, \dots, L-1
\end{equation}
where
\begin{equation}
    \mathrm{CKA}\big(\tilde{\bm{G}}_t^{(l)},\, \tilde{\bm{G}}_t^{(l+1)}\big) = \frac{\mathrm{tr}\big(\tilde{\bm{G}}_t^{(l)}\tilde{\bm{G}}_t^{(l+1)}\big)}{\big\|\tilde{\bm{G}}_t^{(l)}\big\|_F \big\|\tilde{\bm{G}}_t^{(l+1)}\big\|_F}.
\end{equation}
$D_t^{(l)}$ is high when the pairwise token similarity structure reorganizes substantially between layers $l$ and $l+1$, and low when it remains stable. The full depth profile $\bm{D}_t = [D_t^{(1)}, \ldots, D_t^{(L-1)}]$ traces the geometric evolution of step $s_t$ throughout the entire forward pass.

\paragraph{Total Step-Aware Reasoning Energy}
We define the \emph{Step-Aware Reasoning Energy} ($E_t$) as the total accumulated geometric dissimilarity across all layer transitions:
\begin{equation}
    E_t = \sum_{l=1}^{L-1} D_t^{(l)}.
\end{equation}
A high $E_t$ indicates extensive cross-layer reorganization, reflecting that the model expended significant computational effort to refine the internal relational structure of that step. Conversely, a low $E_t$ implies early structural stabilization, characteristic of trivial transitions or routine factual recall.

\subsection{Why CKA over Alternative Measures.}
Several natural alternatives exist for measuring layer-to-layer geometric change, each of which we argue is insufficient for our purpose. 

First, one could directly compare the eigenvalue spectra of $G_t^{(l)}$ and $G_t^{(l+1)}$, inspired by \citet{yu2025explainable}. However, the eigenvectors of each Gram matrix are derived independently at their respective layers, spanning different principal directions with no guaranteed cross-layer correspondence; comparing eigenvalues without aligning their eigenvectors is therefore geometrically meaningless. 

Second, one could compute the Jensen-Shannon Divergence (JSD) between token-level next-token distributions across adjacent layers, as adopted in DoLa \citep{chuang2023dola} and DTR \citep{chen2026think}, and aggregate these scores to the step level by averaging. While this approach is computationally efficient, it treats each token independently and discards all token-token relationship information, precisely the relational structure that the Gram matrix is designed to capture. Since our goal is to measure how the \textit{collective} semantic geometry of a reasoning step evolves across layers, a token-wise measure that ignores inter-token dependencies isfundamentally misaligned with this objective. 

Third, one could compare soft cluster distributions $P(C_m \mid s_t)$ defined in Section~\ref{sec:clustering} via JSD across layers. However, since clusters are derived independently at each layer, there is no guaranteed correspondence between cluster identities: the same cluster index may refer to entirely different semantic groupings at different layers. While this misalignment could in principle be resolved by permuting cluster assignments to find an optimal correspondence, doing so requires solving a combinatorial assignment problem at every layer transition, which is computationally prohibitive at scale. 

CKA circumvents all three limitations by operating directly on the Gram matrices, whose $(i,j)$ entries always refer to the same token pair across all layers by construction. This preserves the full pairwise token relationship structure without requiring any eigenvector alignment, cluster correspondence, or permutation. Furthermore, CKA is invariant to orthogonal transformations and isotropic scaling of hidden states, making it robust to the layer-wise normalization and rotation ambiguities that are common in deep networks.

\section{Experiments}

In this section, we empirically validate our unified framework. By integrating layer-wise geometric dissimilarity with Markovian state transitions, we analyze how reasoning energy varies across semantic roles, evolves over the course of a reasoning trajectory, and whether its patterns can indicate reasoning failure of CoT trajectories independently of the final answer.

\subsection{Data and Models}
We evaluate our framework on six established benchmarks across three reasoning domains, all of which are widely used to assess LLM reasoning ability: (1) \textbf{Math:} \emph{GSM8K} \citep{cobbe2021training} and \emph{MATH} \citep{hendrycks2021measuring}, focusing on grade-school and advanced numerical problem-solving. (2) \textbf{Commonsense:} \emph{CSQA} \citep{talmor2019commonsenseqa} and \emph{StrategyQA} \citep{geva2021did}, challenging the model's intuitive reasoning and implicit factual deductions. (3) \textbf{Multi-Hop:} \emph{HotpotQA} \citep{yang2018hotpotqa} and \emph{MuSiQue} \citep{trivedi2022musique}, requiring multi-step factual inference over textual evidence.

For the LLM backbones, we generate CoT trajectories and extract internal hidden states using three recent, highly-capable models: \textbf{LLaMA-3.2-3B} \citep{grattafiori2024llama}, \textbf{Phi-4-mini} \citep{abouelenin2025phi}, and \textbf{Gemma-3-4B} \citep{gemmateam2025gemma3technicalreport}.

For each dataset we sample 800 examples per model (687 for StrategyQA, where the full test set is used), drawing from the standard test split for GSM8K, MATH, and StrategyQA and the validation split for CSQA, HotpotQA, and MuSiQue. Full protocol details (i.e., decoding parameters, per-dataset answer normalization, and class balance statistics) are provided in Appendix~\ref{sec:protocol}.

\subsection{RQ1: Reasoning Energy Profiles Across Reasoning States}

\begin{figure}[h!]
\centering
\includegraphics[width=.5\textwidth]{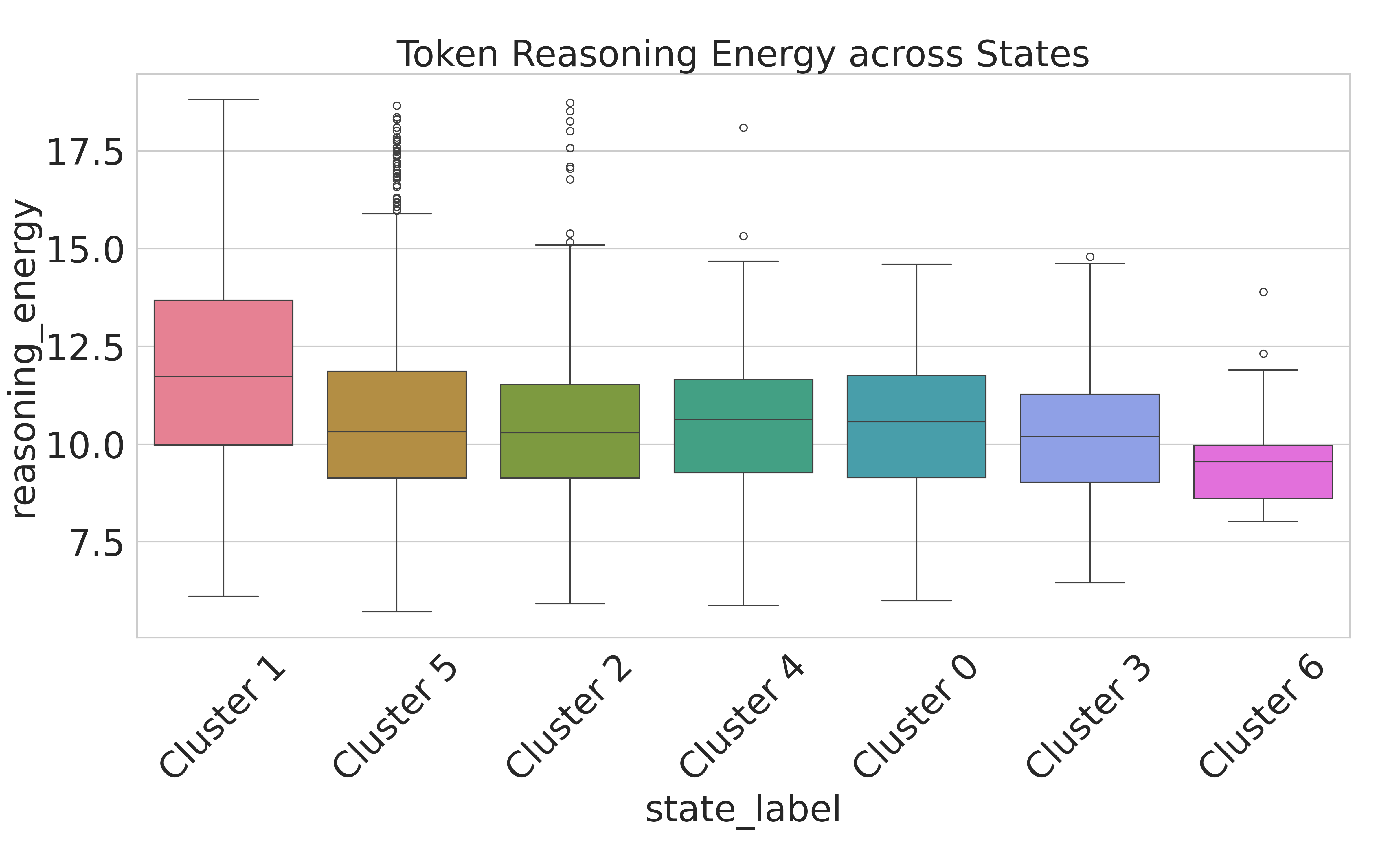}\hfill
\includegraphics[width=.5\textwidth]{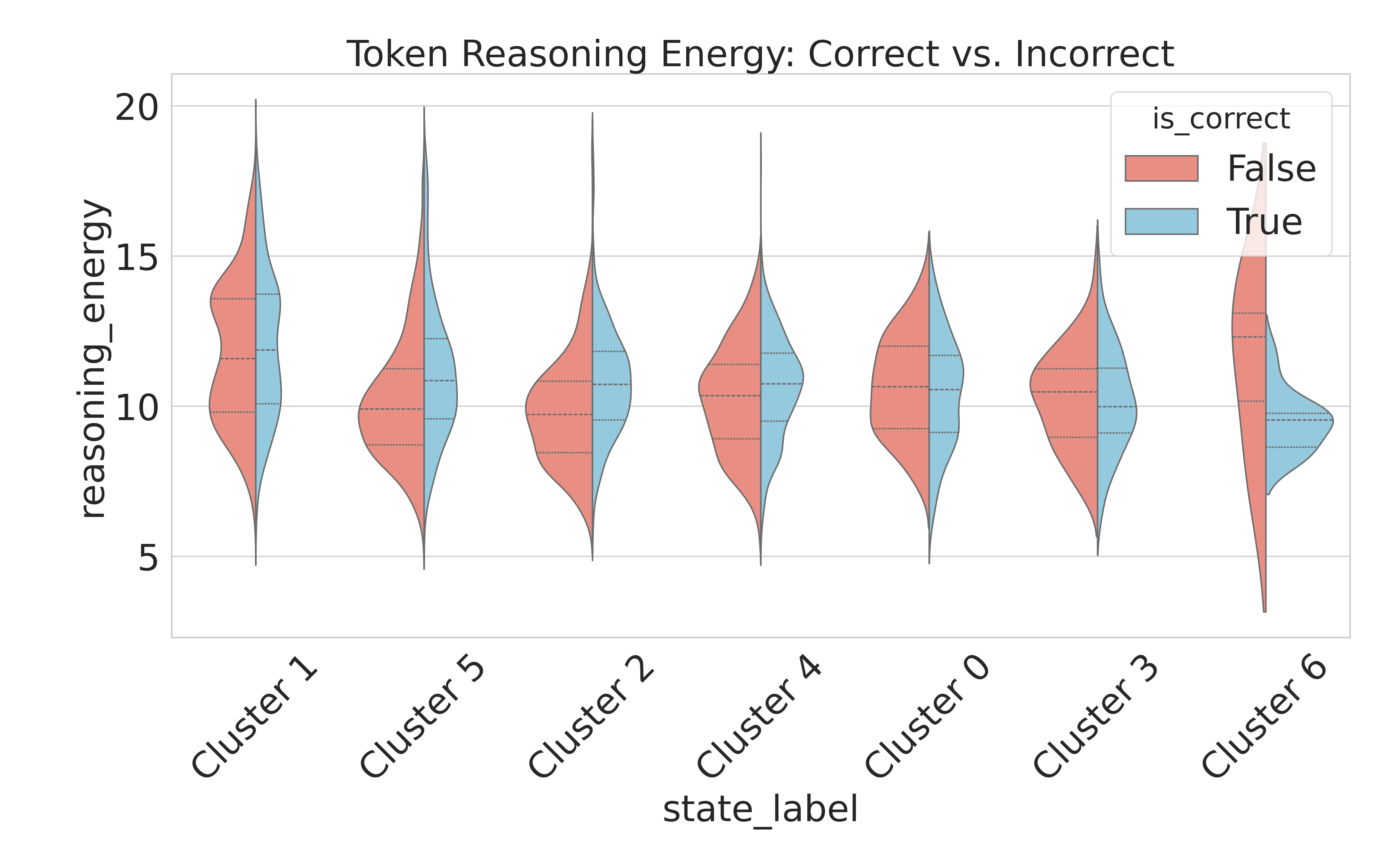}\hfill
\caption{Reasoning energy profiles for each cluster on \textbf{GSM8K} with \textbf{Phi-4-mini}, shown unconditioned (left) and conditioned on trajectory correctness (right).}
\label{fig:rq1_phi_gsm8k}
\end{figure}

\begin{figure}[h!]
\centering
\includegraphics[width=.5\textwidth]{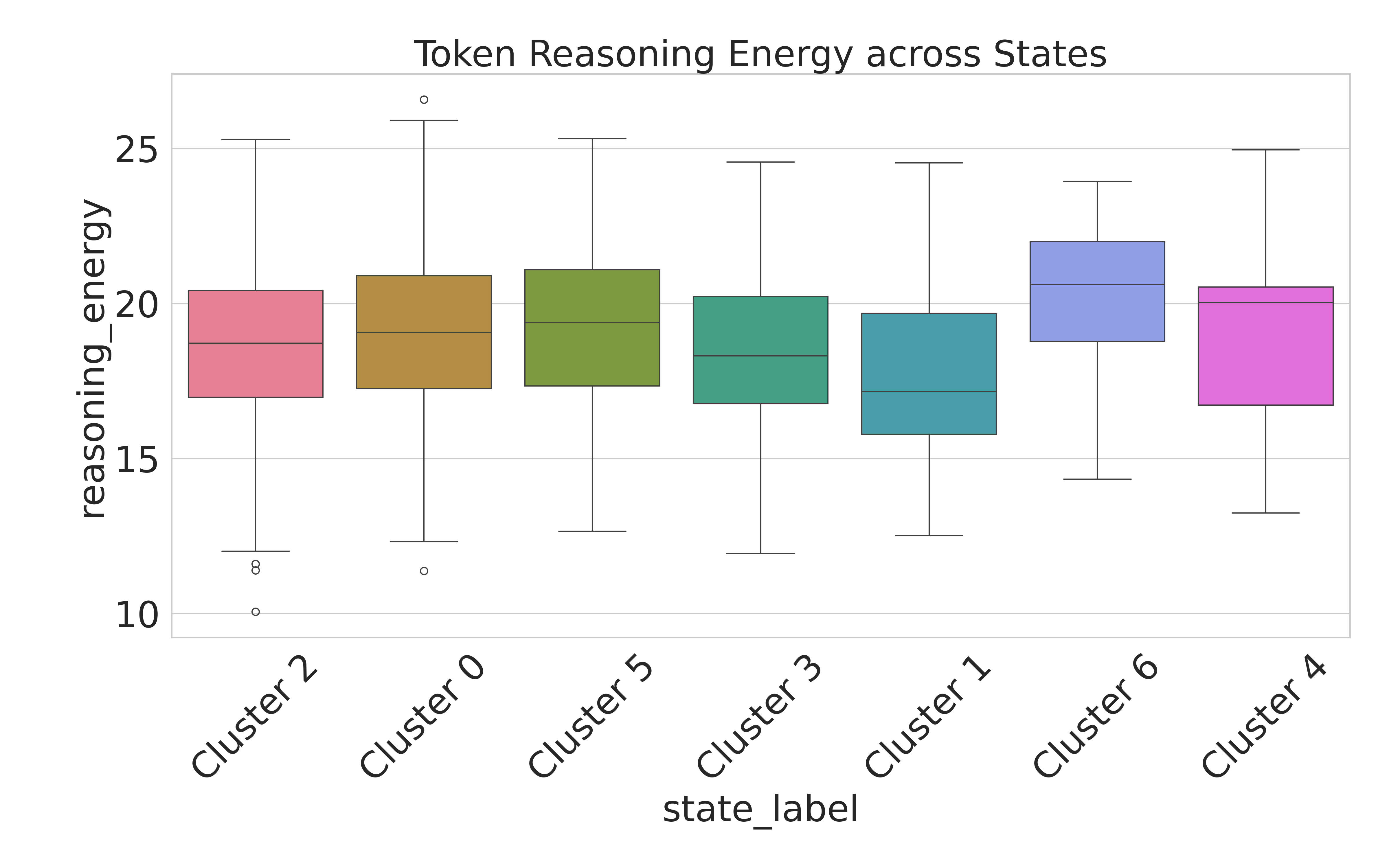}\hfill
\includegraphics[width=.5\textwidth]{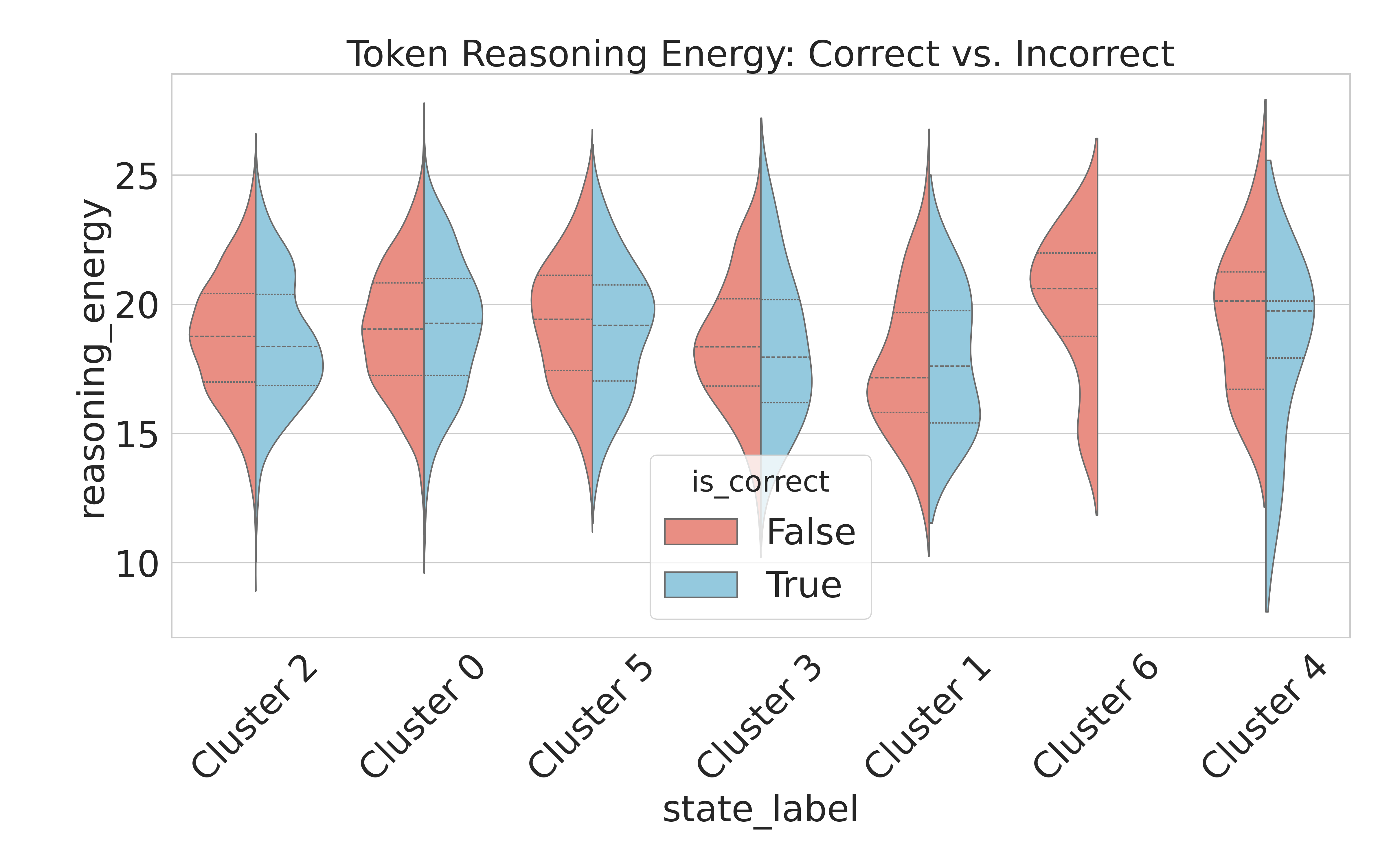}\hfill
\caption{Reasoning energy profiles for each cluster on \textbf{HotpotQA} with \textbf{Gemma-3-4B}, shown unconditioned (left) and conditioned on trajectory correctness (right).}
\label{fig:rq1_gemma_hotpotqa}
\end{figure}

\begin{figure}[h!]
\centering
\includegraphics[width=.5\textwidth]{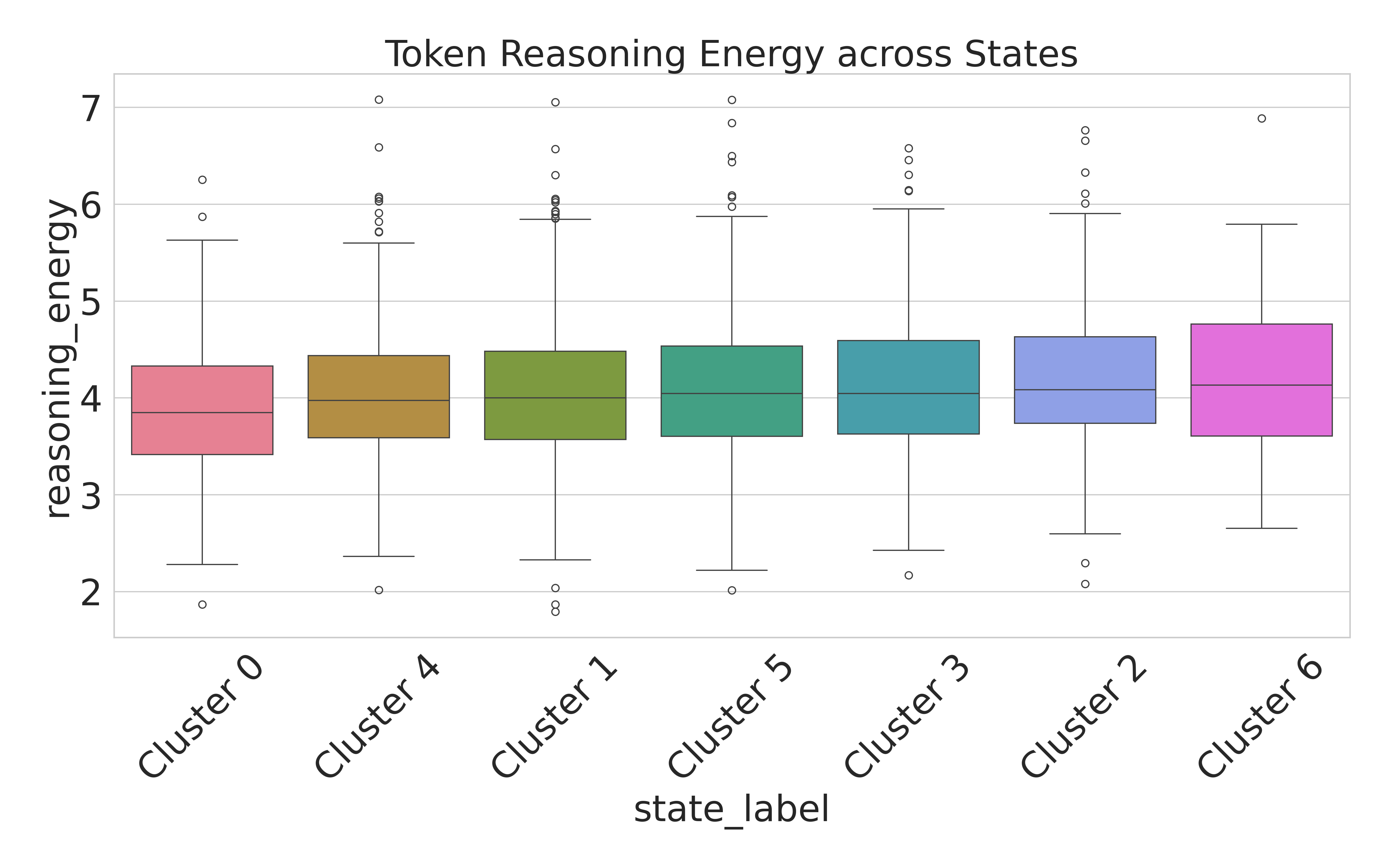}\hfill
\includegraphics[width=.5\textwidth]{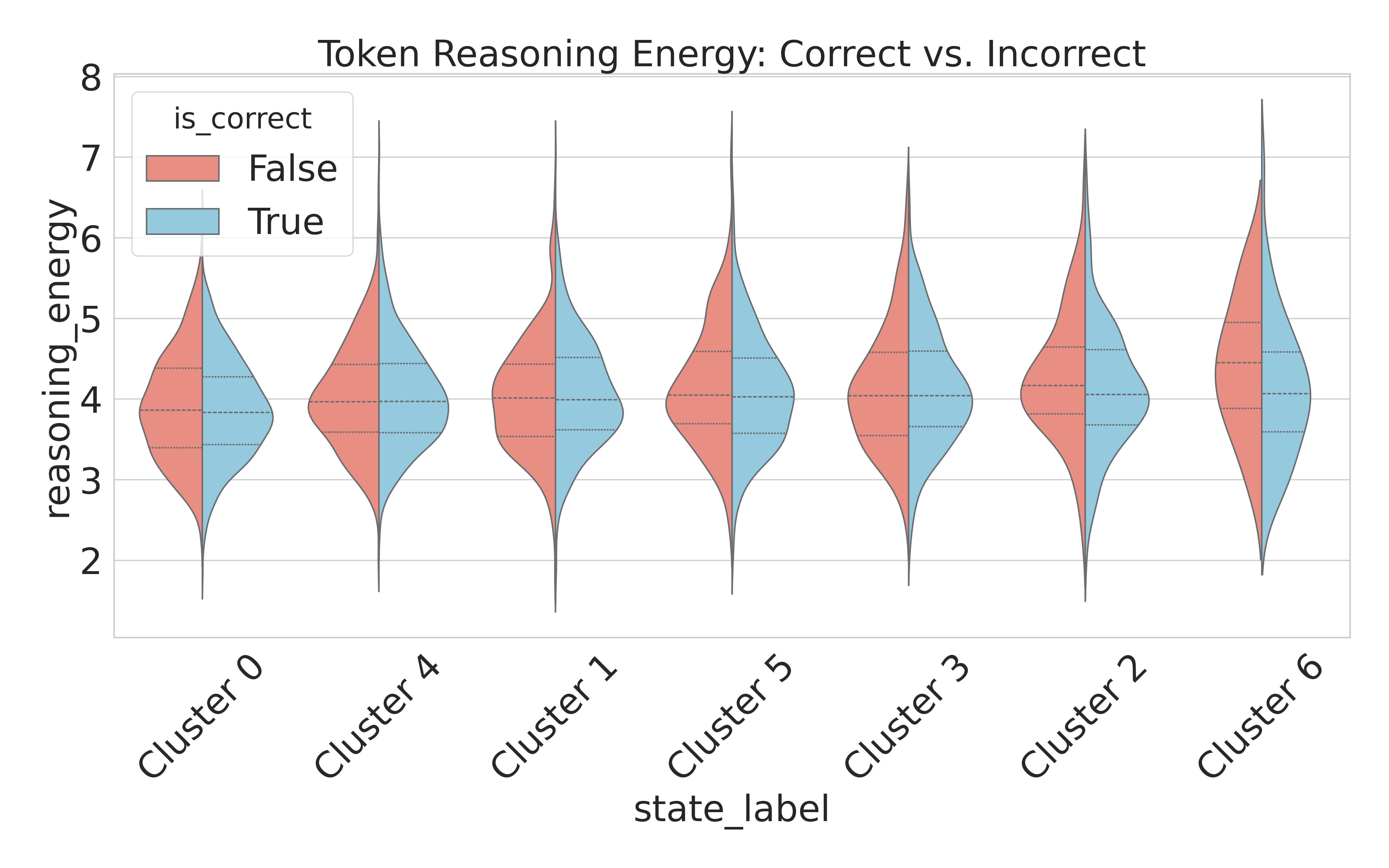}\hfill
\caption{Reasoning energy profiles for each cluster on \textbf{StrategyQA} with \textbf{LLaMA-3.2-3B}, shown unconditioned (left) and conditioned on trajectory correctness (right).}
\label{fig:rq1_llama_strategyqa}
\end{figure}

Our first goal is twofold: (1) to examine whether different reasoning states exhibit distinct step-aware reasoning energy (SARE) profiles, and (2) to investigate whether the energy profiles within each reasoning state differ between correct and incorrect trajectories.

To address these questions, we cluster all reasoning steps generated by the model into $K$ groups ($K = 7$ in our experiments) using the method described in Section~\ref{sec:clustering}, and then compute the step-level reasoning energy defined in Section~\ref{sec:CKA}. To answer the first question, we visualize the SARE distribution of each cluster using \emph{box plots}; to answer the second, we compare the distributions of correct and incorrect trajectories within each cluster using \emph{violin plots}. Figures \ref{fig:rq1_phi_gsm8k} to \ref{fig:rq1_llama_strategyqa} present representative results for Phi-4 on GSM8K, Gemma-3 on HotpotQA and Llama-3.2 on StrategyQA, respectively.  Clusters are arranged according to their most frequently observed positions in the reasoning trajectory.

Our analysis of the SARE profiles across three diverse cognitive domains reveals that internal hidden states are not merely abstract representations, but correspond to distinct functional stages of the reasoning process. By combining the quantitative distributions from box and violin plots with the qualitative examples from our curated clusters, we identify several key findings regarding the energetic cost of CoT generation.

\paragraph{Finding 1: Semantic Interpretation of Reasoning States}
A cross-model examination of the $K=7$ clusters reveals a broadly consistent functional taxonomy of reasoning states. Early-stage clusters typically correspond to \textbf{initial setup and information extraction}, where the model parses the prompt and identifies key variables or entities (e.g., extracting quantities in math problems or identifying relevant entities in multi-hop questions). In \emph{Phi-4-mini} on \emph{GSM8K}, this stage exhibits relatively \emph{high} median energy, suggesting that translating natural language into a structured mathematical representation requires substantial internal transformation. In contrast, for \emph{LLaMA-3.2-3B} on \emph{StrategyQA}, the corresponding early-stage clusters show comparatively \emph{lower} energy than later reasoning states.

Mid-trajectory clusters are generally associated with \textbf{factual retrieval and intermediate comparison}, such as accessing world knowledge or performing local logical inferences. These states tend to exhibit stable, \emph{moderate} energy levels across models. Final-stage clusters typically correspond to \textbf{synthesis and termination}, including integrative reasoning steps and final answer formulation. Notably, in \emph{Gemma-3-4B} and \emph{LLaMA-3.2-3B}, the final cluster often exhibits the \emph{highest} energy, suggesting that the synthesis of disparate evidence is the most computationally demanding stage in multi-hop and commonsense reasoning. By contrast, in \emph{Phi-4-mini}, the final cluster consistently shows the \emph{lowest} energy, indicating that in mathematical reasoning, the last step is often comparatively lightweight once the core reasoning has already been completed.

\vspace{-0.2cm}
\paragraph{Finding 2: Energy-Based Signatures of Reasoning Failure}
The violin plots reveal that incorrect trajectories are associated with distinctive energy patterns: they tend to exhibit lower energy than correct trajectories in specific reasoning states, most notably in final verification or synthesis states, where erroneous trajectories skew toward lower-energy distributions. For example, in Phi-4-mini on GSM8K, incorrect final reasoning steps show a clear shift toward lower SARE. We note that this is a correlational observation: the analysis does not establish whether lower energy precedes or contributes to failure, but rather that the two co-occur at specific reasoning junctions.

This divergence is not limited to the final stage. In LLaMA-3.2, the energy distributions of correct and incorrect trajectories largely overlap in early retrieval states, but begin to diverge in later quantification and compositional reasoning states. This suggests that the correlation between lower energy and incorrect outcomes is localized to specific junctions in the trajectory rather than being uniformly distributed.

\vspace{-0.2cm}
\paragraph{Finding 3: Model-Specific Energetic Profiles}
We observe that the "dynamic range" of reasoning energy is heavily dependent on model architecture. \emph{Gemma-3-4B} utilizes a much wider energy spectrum ($10.0$–$27.5$) than \emph{Llama-3.2-3B} ($2.0$–$7.0$) or \emph{Phi-4-mini} ($6.0$–$18.0$). This implies that larger or differently architectural models may engage in more intensive representational re-writing across transformer layers. 

\subsection{RQ2: Predicting Reasoning Failure via Latent Dynamics}
To evaluate the predictive utility of our internal reasoning signals, we define a binary classification task to detect reasoning errors (incorrect final answers) in an offline setting.

\paragraph{Experimental Design and Evaluation Protocol} \label{sec:experiment_design}
Following common practice in constructing trajectory features \citep{dempster2019rocket, wang2017time}, we aggregate step-aware energy and state information into a 12-dimensional trajectory-level vector consisting of seven \textbf{energy-intensity} statistics (mean, median, standard deviation, range, volatility, peaks, and valleys, where volatility, peaks, and valleys directly operationalize the cross-step energy velocity $\Delta E(s_i \to s_{i+1})$ introduced in Section~3.3 by capturing the magnitude, local maxima, and local minima of step-to-step energy changes) and five \textbf{state-topology} statistics (cluster entropy, state revisits, unique transitions, transition diversity, and most frequent state ratio). We then adopt a \emph{stratified} 70/10/20 train/validation/test split based on final answer correctness. A Logistic Regression model with $\ell_2$ regularization is trained on the 70\% training set, and the classification threshold is tuned on the 10\% validation set to maximize F1-score. The final performance is then evaluated on the held-out 20\% test set.

\paragraph{Baselines}
We compare SARE against five established baselines: (1) \textbf{Token Count} \citep{guo2025deepseek, yang2025qwen3, zhong2024evaluation}: the total length of the CoT trajectory; (2) \textbf{Mean Log-Probability} \citep{kauf2024log, zhang2025cost}: the average log-likelihood of generated tokens; (3) \textbf{Negative Entropy} \citep{zhao2026entropy, buffa2026entropy}: the mean negative Shannon entropy of the token distribution, we report negative entropy since larger values indicate higher confidence; (4) \textbf{Negative Perplexity} \citep{zhou2025bridging, geng2024survey}: calculated as $-\exp(-\text{mean log-likelihood})$, we report negative perplexity since larger values correspond to higher confidence; and (5) \textbf{Self-Certainty} \citep{kang2025scalable}: the model's internal confidence score, computed as the KL-divergence between the token distribution and a uniform distribution. For every baseline, we apply the same threshold-tuning protocol on the validation set to ensure fair comparison.

\paragraph{Reasoning Energy and Token Count as Complementary Signals}
In our main results, we combine the trajectory-level vector described in Section~\ref{sec:experiment_design} with \textbf{Token Count}, as the two provide complementary information and consistently improve performance. In the appendix, we report an ablation study without Token Count. Although absolute performance slightly decreases, the overall trends remain consistent with the results presented here (see Appendix Table \ref{tab:f1_notoken_llama_3_2_3b}-\ref{tab:f1_notoken_phi_4_mini}).

A key design choice of our method is the integration of \textit{Reasoning Energy} and \textit{Token Count}. We view internal energy dynamics and trajectory length as complementary signals of reasoning correctness. While standard probabilistic baselines (e.g., Entropy or Log-Prob) primarily capture final-layer representation properties, which are already implicitly represented in our layer-wise Gram matrices, they often remove the token count information by averaging the token-wise information over the entire trajectory. By incorporating both internal dynamics and overall trajectory length, our method provides a more complete characterization of the model’s reasoning process.

\paragraph{Empirical Findings (F1 Performance):}
Across three LLM families and six datasets (Tables~\ref{tab:f1_full_llama_3_2_3b}--\ref{tab:f1_full_phi_4_mini}), SARE demonstrates competitive discriminative power for reasoning error detection, matching or outperforming most baseline metrics. 

A notable pattern is that four probabilistic baselines (Mean Log-Probability, Negative Entropy, Negative Perplexity, and Self-Certainty) report F1\,=\,0 on StrategyQA across all three models, which is not an implementation error, but a fundamental limitation of output-based confidence for binary True/False tasks, where token-level probability distributions are near-uniform across trajectories regardless of correctness and threshold tuning collapses to predicting all trajectories correct. Neither SARE nor Token Count exhibits this failure, retaining F1 of 0.46--0.53 and 0.46--0.58 respectively, as both operate on full trajectory features rather than final-token probabilities. 

We note, however, that Token Count matches or outperforms SARE on StrategyQA for LLaMA-3.2-3B and Gemma-3-4B, suggesting that for short binary-answer tasks trajectory length captures much of the available signal; comparisons with the collapsed baselines on this dataset should be interpreted accordingly. This discriminative advantage over probabilistic baselines persists when trajectory length is excluded from the feature set, indicating that internal geometric dynamics encode predictive information beyond surface-level heuristics. AUPRC results (Appendix Tables~\ref{tab:auprc_full_llama_3_2_3b}--\ref{tab:auprc_full_phi_4_mini}) confirm the same overall trends.

\begin{table*}[h!]
\centering
\small
\caption{F1 Comparison: SARE vs. All Baselines for \textbf{LLaMA-3.2-3B}}
\label{tab:f1_full_llama_3_2_3b}
\begin{tabular*}{\linewidth}{@{\extracolsep{\fill}}lc|ccccc}
\toprule
Dataset & SARE & Token & LogProb & NegEnt & NegPerp & SelfCert \\
\midrule
GSM8K      & \textbf{0.532} & 0.514 & 0.512 & 0.507 & 0.512 & 0.505 \\
MATH       & \textbf{0.801} & 0.789 & 0.792 & 0.792 & 0.792 & 0.792 \\
CSQA       & \textbf{0.454} & 0.448 & 0.436 & 0.431 & 0.436 & 0.423 \\
HotpotQA   & \textbf{0.900} & 0.900 & \textbf{0.900} & \textbf{0.900} & \textbf{0.900} & \textbf{0.900} \\
MuSiQue    & \textbf{0.958} & 0.947 & 0.954 & 0.954 & 0.954 & 0.947 \\
StrategyQA & 0.491 & \textbf{0.580} & 0.000 & 0.000 & 0.000 & 0.000 \\
\bottomrule
\end{tabular*}
\end{table*}

\begin{table*}[h!]
\centering
\small
\caption{F1 Comparison: SARE vs. All Baselines for \textbf{Gemma-3-4B}}
\label{tab:f1_full_gemma_3_4b}
\begin{tabular*}{\linewidth}{@{\extracolsep{\fill}}lc|ccccc}
\toprule
Dataset & SARE & Token & LogProb & NegEnt & NegPerp & SelfCert \\
\midrule
GSM8K      & \textbf{0.293} & 0.200 & 0.237 & 0.262 & 0.236 & 0.282 \\
MATH       & \textbf{0.746} & 0.691 & 0.640 & 0.634 & 0.640 & 0.673 \\
CSQA       & \textbf{0.473} & 0.471 & 0.451 & 0.415 & 0.451 & 0.465 \\
HotpotQA   & \textbf{0.893} & 0.885 & \textbf{0.893} & 0.864 & \textbf{0.893} & 0.889 \\
MuSiQue    & \textbf{0.964} & \textbf{0.964} & 0.961 & \textbf{0.964} & 0.961 & \textbf{0.964} \\
StrategyQA & 0.459 & \textbf{0.460} & 0.000 & 0.000 & 0.000 & 0.000 \\
\bottomrule
\end{tabular*}
\end{table*}

\begin{table*}[h!]
\centering
\small
\caption{F1 Comparison: SARE vs. All Baselines for \textbf{Phi-4-Mini}}
\label{tab:f1_full_phi_4_mini}
\begin{tabular*}{\linewidth}{@{\extracolsep{\fill}}lc|ccccc}
\toprule
Dataset & SARE & Token & LogProb & NegEnt & NegPerp & SelfCert \\
\midrule
GSM8K      & \textbf{0.621} & 0.617 & 0.615 & 0.609 & 0.615 & 0.615 \\
MATH       & \textbf{0.769} & \textbf{0.769} & 0.755 & 0.764 & 0.755 & 0.755 \\
CSQA       & 0.462 & \textbf{0.464} & 0.462 & 0.462 & 0.462 & \textbf{0.464} \\
HotpotQA   & \textbf{0.908} & 0.907 & 0.904 & \textbf{0.908} & 0.904 & 0.891 \\
MuSiQue    & \textbf{0.974} & 0.968 & \textbf{0.974} & \textbf{0.974} & \textbf{0.974} & 0.968 \\
StrategyQA & \textbf{0.526} & 0.514 & 0.000 & 0.000 & 0.000 & 0.000 \\
\bottomrule
\end{tabular*}
\end{table*}

\section{Conclusion}

In this paper, we proposed \textbf{Step-Aware Reasoning Energy (SARE)}, a
geometric framework that quantifies computational effort at the level of
individual reasoning steps in chain-of-thought trajectories. By applying
Centered Kernel Alignment (CKA) to token hidden state Gram matrices across
consecutive transformer layers, SARE measures how much each step's internal
token relationship geometry reorganizes during the forward pass, without
requiring eigenvector alignment, cluster correspondence, or token-level
aggregation. Experiments across six reasoning benchmarks and three open-source
LLMs reveal that reasoning energy is highly non-uniform across semantic step
types, that incorrect trajectories exhibit systematically lower energy at
critical reasoning junctions prior to failure, and that energy dynamics encode
predictive information about reasoning correctness beyond surface-level
confidence measures.

In its current form, SARE is primarily a diagnostic and interpretability tool:
it reveals \emph{where} and \emph{how much} computational effort is expended
across a reasoning trajectory, and identifies energetic signatures that
correlate with incorrect outcomes. Translating these diagnostic signals into
actionable interventions is a natural next step. SARE scores could serve as a
training-time signal to encourage the model to sustain higher geometric
reorganization at critical reasoning junctions, analogous to process-reward
models that supervise intermediate steps. At inference
time, low-energy steps at high-stakes junctions could trigger targeted
re-sampling or beam search expansion, and SARE profiles could inform early-exit
strategies for steps that stabilize quickly. We hope the structured energy
patterns identified here provide a useful foundation for geometry-aware
reasoning improvement.
\section*{Limitations}

\paragraph{Model scale.} All experiments use open-weight LLMs in the 3--4B parameter range, spanning three architecturally distinct families (LLaMA-3.2-3B, Phi-4-mini, Gemma-3-4B). Whether the observed energy profiles and failure-detection patterns generalize to larger models is an open empirical question we leave to future work. That said, the consistency of findings across three families differing in training objectives, tokenizers, and attention configurations already provides initial cross-architecture evidence that the geometric signal captured by CKA is not model-specific, offering some basis for expecting broader generalizability.

\paragraph{Computational overhead.} SARE requires extracting hidden states at every transformer layer for every reasoning step, making it best suited to offline analysis in its current form. Approximations such as layer subsampling or mini-batch CKA estimation are natural directions for reducing this cost in large-scale settings.

\paragraph{LLM Usage.} An LLM coding assistant was used to support portions of the implementation.

\bibliography{custom}

@inproceedings{yu2025explainable,
  author       = {Sheldon Yu and
                  Yuxin Xiong and
                  Junda Wu and
                  Xintong Li and
                  Tong Yu and
                  Xiang Chen and
                  Ritwik Sinha and
                  Jingbo Shang and
                  Julian J. McAuley},
  
  title        = {Explainable Chain-of-Thought Reasoning: An Empirical Analysis on State-Aware
                  Reasoning Dynamics},
  booktitle    = {Findings of the Association for Computational Linguistics: {EMNLP}
                  2025, Suzhou, China, November 4-9, 2025},
  pages        = {16660--16667},
  publisher    = {Association for Computational Linguistics},
  year         = {2025},
  url          = {https://aclanthology.org/2025.findings-emnlp.904/},
  bibsource    = {dblp computer science bibliography, https://dblp.org}
}

@article{chen2026think,
  title={Think Deep, Not Just Long: Measuring LLM Reasoning Effort via Deep-Thinking Tokens},
  author={Chen, Wei-Lin and Peng, Liqian and Tan, Tian and Zhao, Chao and Chen, Blake JianHang and Lin, Ziqian and Go, Alec and Meng, Yu},
  journal={arXiv preprint arXiv:2602.13517},
  year={2026}
}

@article{wei2022chain,
  title={Chain-of-thought prompting elicits reasoning in large language models},
  author={Wei, Jason and Wang, Xuezhi and Schuurmans, Dale and Bosma, Maarten and Xia, Fei and Chi, Ed and Le, Quoc V and Zhou, Denny and others},
  journal={Advances in neural information processing systems},
  volume={35},
  pages={24824--24837},
  year={2022}
}

@article{wang2022self,
  title={Self-consistency improves chain of thought reasoning in language models},
  author={Wang, Xuezhi and Wei, Jason and Schuurmans, Dale and Le, Quoc and Chi, Ed and Narang, Sharan and Chowdhery, Aakanksha and Zhou, Denny},
  journal={arXiv preprint arXiv:2203.11171},
  year={2022}
}

@article{hwang2026oops,
  title={Oops, Wait: Token-Level Signals as a Lens into LLM Reasoning},
  author={Hwang, Jaehui and Han, Dongyoon and Yun, Sangdoo and Heo, Byeongho},
  journal={arXiv preprint arXiv:2601.17421},
  year={2026}
}

@article{madaan2023self,
  title={Self-refine: Iterative refinement with self-feedback},
  author={Madaan, Aman and Tandon, Niket and Gupta, Prakhar and Hallinan, Skyler and Gao, Luyu and Wiegreffe, Sarah and Alon, Uri and Dziri, Nouha and Prabhumoye, Shrimai and Yang, Yiming and others},
  journal={Advances in neural information processing systems},
  volume={36},
  pages={46534--46594},
  year={2023}
}

@article{belrose2023eliciting,
  title={Eliciting latent predictions from transformers with the tuned lens},
  author={Belrose, Nora and Furman, Zach and Smith, Logan and Halawi, Danny and Ostrovsky, Igor and McKinney, Lev and Biderman, Stella and Steinhardt, Jacob},
  journal={arXiv preprint arXiv:2303.08112},
  year={2023}
}

@article{chuang2023dola,
  title={Dola: Decoding by contrasting layers improves factuality in large language models},
  author={Chuang, Yung-Sung and Xie, Yujia and Luo, Hongyin and Kim, Yoon and Glass, James and He, Pengcheng},
  journal={arXiv preprint arXiv:2309.03883},
  year={2023}
}

@inproceedings{kornblith2019similarity,
  title={Similarity of neural network representations revisited},
  author={Kornblith, Simon and Norouzi, Mohammad and Lee, Honglak and Hinton, Geoffrey},
  booktitle={International conference on machine learning},
  pages={3519--3529},
  year={2019},
  organization={PMlR}
}

@inproceedings{schwartz2020right,
  title={The right tool for the job: Matching model and instance complexities},
  author={Schwartz, Roy and Stanovsky, Gabriel and Swayamdipta, Swabha and Dodge, Jesse and Smith, Noah A},
  booktitle={Proceedings of the 58th Annual Meeting of the Association for Computational Linguistics},
  pages={6640--6651},
  year={2020}
}

@inproceedings{teerapittayanon2016branchynet,
  title={Branchynet: Fast inference via early exiting from deep neural networks},
  author={Teerapittayanon, Surat and McDanel, Bradley and Kung, Hsiang-Tsung},
  booktitle={2016 23rd international conference on pattern recognition (ICPR)},
  pages={2464--2469},
  year={2016},
  organization={IEEE}
}

@article{cobbe2021training,
  title={Training verifiers to solve math word problems},
  author={Cobbe, Karl and Kosaraju, Vineet and Bavarian, Mohammad and Chen, Mark and Jun, Heewoo and Kaiser, Lukasz and Plappert, Matthias and Tworek, Jerry and Hilton, Jacob and Nakano, Reiichiro and others},
  journal={arXiv preprint arXiv:2110.14168},
  year={2021}
}

@article{hendrycks2021measuring,
  title={Measuring mathematical problem solving with the math dataset},
  author={Hendrycks, Dan and Burns, Collin and Kadavath, Saurav and Arora, Akul and Basart, Steven and Tang, Eric and Song, Dawn and Steinhardt, Jacob},
  journal={arXiv preprint arXiv:2103.03874},
  year={2021}
}

@inproceedings{talmor2019commonsenseqa,
  title={Commonsenseqa: A question answering challenge targeting commonsense knowledge},
  author={Talmor, Alon and Herzig, Jonathan and Lourie, Nicholas and Berant, Jonathan},
  booktitle={Proceedings of the 2019 Conference of the North American Chapter of the Association for Computational Linguistics: Human Language Technologies, Volume 1 (Long and Short Papers)},
  pages={4149--4158},
  year={2019}
}

@article{geva2021did,
  title={Did aristotle use a laptop? a question answering benchmark with implicit reasoning strategies},
  author={Geva, Mor and Khashabi, Daniel and Segal, Elad and Khot, Tushar and Roth, Dan and Berant, Jonathan},
  journal={Transactions of the Association for Computational Linguistics},
  volume={9},
  pages={346--361},
  year={2021},
  publisher={MIT Press One Rogers Street, Cambridge, MA 02142-1209, USA journals-info~…}
}

@inproceedings{yang2018hotpotqa,
  title={HotpotQA: A dataset for diverse, explainable multi-hop question answering},
  author={Yang, Zhilin and Qi, Peng and Zhang, Saizheng and Bengio, Yoshua and Cohen, William and Salakhutdinov, Ruslan and Manning, Christopher D},
  booktitle={Proceedings of the 2018 conference on empirical methods in natural language processing},
  pages={2369--2380},
  year={2018}
}

@article{trivedi2022musique,
  title={MuSiQue: Multihop Questions via Single-hop Question Composition},
  author={Trivedi, Harsh and Balasubramanian, Niranjan and Khot, Tushar and Sabharwal, Ashish},
  journal={Transactions of the Association for Computational Linguistics},
  volume={10},
  pages={539--554},
  year={2022},
  publisher={MIT Press One Broadway, 12th Floor, Cambridge, Massachusetts 02142, USA~…}
}

@article{grattafiori2024llama,
  title={The llama 3 herd of models},
  author={Grattafiori, Aaron and Dubey, Abhimanyu and Jauhri, Abhinav and Pandey, Abhinav and Kadian, Abhishek and Al-Dahle, Ahmad and Letman, Aiesha and Mathur, Akhil and Schelten, Alan and Vaughan, Alex and others},
  journal={arXiv preprint arXiv:2407.21783},
  year={2024}
}

@article{abouelenin2025phi,
  title={Phi-4-mini technical report: Compact yet powerful multimodal language models via mixture-of-loras},
  author={Abouelenin, Abdelrahman and Ashfaq, Atabak and Atkinson, Adam and Awadalla, Hany and Bach, Nguyen and Bao, Jianmin and Benhaim, Alon and Cai, Martin and Chaudhary, Vishrav and Chen, Congcong and others},
  journal={arXiv preprint arXiv:2503.01743},
  year={2025}
}

@misc{gemmateam2025gemma3technicalreport,
      title={Gemma 3 Technical Report}, 
      author={Gemma Team and Aishwarya Kamath and Johan Ferret and Shreya Pathak and Nino Vieillard and Ramona Merhej and Sarah Perrin and Tatiana Matejovicova and Alexandre Ramé and Morgane Rivière and Louis Rouillard and Thomas Mesnard and Geoffrey Cideron and Jean-bastien Grill and Sabela Ramos and Edouard Yvinec and Michelle Casbon and Etienne Pot and Ivo Penchev and Gaël Liu and Francesco Visin and Kathleen Kenealy and Lucas Beyer and Xiaohai Zhai and Anton Tsitsulin and Robert Busa-Fekete and Alex Feng and Noveen Sachdeva and Benjamin Coleman and Yi Gao and Basil Mustafa and Iain Barr and Emilio Parisotto and David Tian and Matan Eyal and Colin Cherry and Jan-Thorsten Peter and Danila Sinopalnikov and Surya Bhupatiraju and Rishabh Agarwal and Mehran Kazemi and Dan Malkin and Ravin Kumar and David Vilar and Idan Brusilovsky and Jiaming Luo and Andreas Steiner and Abe Friesen and Abhanshu Sharma and Abheesht Sharma and Adi Mayrav Gilady and Adrian Goedeckemeyer and Alaa Saade and Alex Feng and Alexander Kolesnikov and Alexei Bendebury and Alvin Abdagic and Amit Vadi and András György and André Susano Pinto and Anil Das and Ankur Bapna and Antoine Miech and Antoine Yang and Antonia Paterson and Ashish Shenoy and Ayan Chakrabarti and Bilal Piot and Bo Wu and Bobak Shahriari and Bryce Petrini and Charlie Chen and Charline Le Lan and Christopher A. Choquette-Choo and CJ Carey and Cormac Brick and Daniel Deutsch and Danielle Eisenbud and Dee Cattle and Derek Cheng and Dimitris Paparas and Divyashree Shivakumar Sreepathihalli and Doug Reid and Dustin Tran and Dustin Zelle and Eric Noland and Erwin Huizenga and Eugene Kharitonov and Frederick Liu and Gagik Amirkhanyan and Glenn Cameron and Hadi Hashemi and Hanna Klimczak-Plucińska and Harman Singh and Harsh Mehta and Harshal Tushar Lehri and Hussein Hazimeh and Ian Ballantyne and Idan Szpektor and Ivan Nardini and Jean Pouget-Abadie and Jetha Chan and Joe Stanton and John Wieting and Jonathan Lai and Jordi Orbay and Joseph Fernandez and Josh Newlan and Ju-yeong Ji and Jyotinder Singh and Kat Black and Kathy Yu and Kevin Hui and Kiran Vodrahalli and Klaus Greff and Linhai Qiu and Marcella Valentine and Marina Coelho and Marvin Ritter and Matt Hoffman and Matthew Watson and Mayank Chaturvedi and Michael Moynihan and Min Ma and Nabila Babar and Natasha Noy and Nathan Byrd and Nick Roy and Nikola Momchev and Nilay Chauhan and Noveen Sachdeva and Oskar Bunyan and Pankil Botarda and Paul Caron and Paul Kishan Rubenstein and Phil Culliton and Philipp Schmid and Pier Giuseppe Sessa and Pingmei Xu and Piotr Stanczyk and Pouya Tafti and Rakesh Shivanna and Renjie Wu and Renke Pan and Reza Rokni and Rob Willoughby and Rohith Vallu and Ryan Mullins and Sammy Jerome and Sara Smoot and Sertan Girgin and Shariq Iqbal and Shashir Reddy and Shruti Sheth and Siim Põder and Sijal Bhatnagar and Sindhu Raghuram Panyam and Sivan Eiger and Susan Zhang and Tianqi Liu and Trevor Yacovone and Tyler Liechty and Uday Kalra and Utku Evci and Vedant Misra and Vincent Roseberry and Vlad Feinberg and Vlad Kolesnikov and Woohyun Han and Woosuk Kwon and Xi Chen and Yinlam Chow and Yuvein Zhu and Zichuan Wei and Zoltan Egyed and Victor Cotruta and Minh Giang and Phoebe Kirk and Anand Rao and Kat Black and Nabila Babar and Jessica Lo and Erica Moreira and Luiz Gustavo Martins and Omar Sanseviero and Lucas Gonzalez and Zach Gleicher and Tris Warkentin and Vahab Mirrokni and Evan Senter and Eli Collins and Joelle Barral and Zoubin Ghahramani and Raia Hadsell and Yossi Matias and D. Sculley and Slav Petrov and Noah Fiedel and Noam Shazeer and Oriol Vinyals and Jeff Dean and Demis Hassabis and Koray Kavukcuoglu and Clement Farabet and Elena Buchatskaya and Jean-Baptiste Alayrac and Rohan Anil and Dmitry and Lepikhin and Sebastian Borgeaud and Olivier Bachem and Armand Joulin and Alek Andreev and Cassidy Hardin and Robert Dadashi and Léonard Hussenot},
      year={2025},
      eprint={2503.19786},
      archivePrefix={arXiv},
      primaryClass={cs.CL},
      url={https://arxiv.org/abs/2503.19786}, 
}

@article{guo2025deepseek,
  title={Deepseek-r1: Incentivizing reasoning capability in llms via reinforcement learning},
  author={Guo, Daya and Yang, Dejian and Zhang, Haowei and Song, Junxiao and Wang, Peiyi and Zhu, Qihao and Xu, Runxin and Zhang, Ruoyu and Ma, Shirong and Bi, Xiao and others},
  journal={arXiv preprint arXiv:2501.12948},
  year={2025}
}

@article{yang2025qwen3,
  title={Qwen3 technical report},
  author={Yang, An and Li, Anfeng and Yang, Baosong and Zhang, Beichen and Hui, Binyuan and Zheng, Bo and Yu, Bowen and Gao, Chang and Huang, Chengen and Lv, Chenxu and others},
  journal={arXiv preprint arXiv:2505.09388},
  year={2025}
}

@article{zhong2024evaluation,
  title={Evaluation of openai o1: Opportunities and challenges of agi},
  author={Zhong, Tianyang and Liu, Zhengliang and Pan, Yi and Zhang, Yutong and Zhang, Zeyu and Zhou, Yifan and Liang, Shizhe and Wu, Zihao and Lyu, Yanjun and Shu, Peng and others},
  journal={arXiv preprint arXiv:2409.18486},
  year={2024}
}

@article{kang2025scalable,
  title={Scalable best-of-n selection for large language models via self-certainty},
  author={Kang, Zhewei and Zhao, Xuandong and Song, Dawn},
  journal={arXiv preprint arXiv:2502.18581},
  year={2025}
}

@article{zhang2025cost,
  title={Cost-augmented monte carlo tree search for llm-assisted planning},
  author={Zhang, Zihao and Liu, Fei},
  journal={arXiv preprint arXiv:2505.14656},
  year={2025}
}

@inproceedings{kauf2024log,
  title={Log probabilities are a reliable estimate of semantic plausibility in base and instruction-tuned language models},
  author={Kauf, Carina and Chersoni, Emmanuele and Lenci, Alessandro and Fedorenko, Evelina and Ivanova, Anna A},
  booktitle={Proceedings of the 7th BlackboxNLP Workshop: Analyzing and Interpreting Neural Networks for NLP},
  pages={263--277},
  year={2024}
}

@article{zhao2026entropy,
  title={Entropy trajectory shape predicts LLM reasoning reliability: A diagnostic study of uncertainty dynamics in chain-of-thought},
  author={Zhao, Xinghao},
  journal={arXiv preprint arXiv:2603.18940},
  year={2026}
}

@article{buffa2026entropy,
  title={Entropy Sentinel: Continuous LLM Accuracy Monitoring from Decoding Entropy Traces in STEM},
  author={Buffa, Pedro Memoli and Del Corro, Luciano},
  journal={arXiv preprint arXiv:2601.09001},
  year={2026}
}

@article{zhou2025bridging,
  title={Bridging internal probability and self-consistency for effective and efficient llm reasoning},
  author={Zhou, Zhi and Yuhao, Tan and Li, Zenan and Yao, Yuan and Guo, Lan-Zhe and Ma, Xiaoxing and Li, Yu-Feng},
  journal={arXiv preprint arXiv:2502.00511},
  year={2025}
}

@inproceedings{geng2024survey,
  title={A survey of confidence estimation and calibration in large language models},
  author={Geng, Jiahui and Cai, Fengyu and Wang, Yuxia and Koeppl, Heinz and Nakov, Preslav and Gurevych, Iryna},
  booktitle={Proceedings of the 2024 Conference of the North American Chapter of the Association for Computational Linguistics: Human Language Technologies (Volume 1: Long Papers)},
  pages={6577--6595},
  year={2024}
}

@article{dempster2019rocket,
  title={ROCKET: exceptionally fast and accurate time series classification using random convolutional kernels},
  author={Dempster, Angus and Petitjean, Fran{\c{c}}ois and Webb, Geoffrey I},
  journal={arXiv preprint arXiv:1910.13051},
  year={2019}
}

@inproceedings{wang2017time,
  title={Time series classification from scratch with deep neural networks: A strong baseline},
  author={Wang, Zhiguang and Yan, Weizhong and Oates, Tim},
  booktitle={2017 International joint conference on neural networks (IJCNN)},
  pages={1578--1585},
  year={2017},
  organization={IEEE}
}

\appendix

\newpage
\appendix
\section{Appendix}

\subsection{Semantic State Clustering: Implementation Details}
\label{sec:clustering_details}

Following \citet{yu2025explainable}, we construct a spectral embedding for each reasoning step in three stages.

\paragraph{Step segmentation.}
CoT trajectories are generated using structured prompts (see Appendix~\ref{sec:prompts}) that instruct the model to begin each reasoning step with the delimiter \texttt{Step X:} on its own line. Steps are extracted by matching this header pattern via regular expression.

\paragraph{Spectral embedding.}
For each sample, the full prompt-plus-trajectory text is passed through the LLM to obtain last-layer token hidden states. The hidden states are projected to 128 dimensions via a fixed random linear projection. For step $s_t$, the corresponding token span is identified via character-offset mapping, and a cumulative feature-covariance matrix is computed as $M_t = \sum_{i=1}^{t} \bm{H}_i^\top \bm{H}_i \in \mathbb{R}^{128 \times 128}$, where $\bm{H}_i$ is the projected token embedding matrix of step $s_i$. The top-64 eigenvalues of $M_t$ by magnitude, computed via sparse eigendecomposition, form the 64-dimensional spectral embedding of $s_t$.

\paragraph{Clustering.}
K-Means clustering ($K{=}7$, Euclidean distance, \texttt{n\_init=10}, \texttt{random\_state=42}) is applied to the step embeddings pooled across all steps and samples in the dataset, yielding a hard cluster label for each step. The cluster sequence of a trajectory defines its state sequence for the Markov transition model.

\subsection{Prompt Templates for CoT Generation}
\label{sec:prompts}

All CoT trajectories are generated using structured prompts that enforce explicit step delimiters.
Each model is instructed to begin every reasoning step with \texttt{Step X:} on its own line,
and to conclude with \texttt{The final answer is: \{answer\}}.
Steps are subsequently extracted by matching this header pattern via regular expression
\texttt{(Step\textbackslash{}s*\textbackslash{}d+\textbackslash{}s*:.*?)(?=Step\textbackslash{}s*\textbackslash{}d+\textbackslash{}s*:|{\$})}.
Below we show representative prompts for each reasoning domain.
Prompts for LLaMA-3.2-3B and Phi-4-mini are identical in format;
Gemma-3-4B prompts additionally include a one-shot worked example to improve format compliance.

\paragraph{Math (LLaMA-3.2-3B / Phi-4-mini, e.g.\ GSM8K).}
\begin{mdframed}
\small
\texttt{Solve the problem step-by-step. Your response must follow these strict format rules:}\\
\texttt{1. Start your response directly with `Step 1:'. Do NOT include any introductory text or pleasantries.}\\
\texttt{2. Each logical step must begin with `Step X: ' (where X is the sequential number).}\\
\texttt{3. Each step must be on a new line.}\\
\texttt{4. Each step must be on a single line.}\\
\texttt{5. End the entire solution with the phrase: `The final answer is: \{number\}'.}\\
\texttt{6. Do not include any additional text or explanation after the final answer.}\\
\texttt{7. Do not repeat the problem or your response once you get the first final answer.}\\[4pt]
\texttt{Problem: \{question\}}\\[4pt]
\texttt{Solution:}
\end{mdframed}

\paragraph{Math (Gemma-3-4B, e.g.\ GSM8K) — includes one-shot example.}
\begin{mdframed}
\small
\texttt{Solve the problem step-by-step. Keep your reasoning concise and to the point.}\\
\texttt{[same format rules 1--4 as above]}\\
\texttt{5. End the entire solution with the phrase: `The final answer is: \{number\}'.}\\
\texttt{[rules 6--8: no extra text, no repeated output, stop after final answer]}\\[4pt]
\texttt{--- Example ---}\\
\texttt{Problem: If John has 5 apples and eats 2, how many does he have left?}\\
\texttt{Solution:}\\
\texttt{Step 1: Identify the initial number of apples, which is 5.}\\
\texttt{Step 2: Identify the number of apples eaten, which is 2.}\\
\texttt{Step 3: Subtract the number of apples eaten from the initial number: 5 - 2 = 3.}\\
\texttt{The final answer is: 3}\\
\texttt{--- End of Example ---}\\[4pt]
\texttt{--- Task ---}\\
\texttt{Problem: \{question\}}\\[4pt]
\texttt{Solution:}
\end{mdframed}

\paragraph{Multi-hop QA (LLaMA-3.2-3B / Phi-4-mini, e.g.\ HotpotQA / MuSiQue).}
\begin{mdframed}
\small
\texttt{Solve the multi-hop reasoning question step-by-step. Your response must follow these strict format rules:}\\
\texttt{[same format rules 1--4]}\\
\texttt{5. End the entire solution with the phrase: `The final answer is: [Answer]'}\\
\texttt{\ \ \ where [Answer] is your short final result.}\\
\texttt{[rules 6--7: no extra text, no repeated output]}\\[4pt]
\texttt{Question: \{question\}}\\[4pt]
\texttt{Solution:}
\end{mdframed}

\paragraph{Commonsense QA (LLaMA-3.2-3B / Phi-4-mini, e.g.\ CSQA / StrategyQA).}
\begin{mdframed}
\small
\texttt{Solve the commonsense question step-by-step using the provided options.}\\
\texttt{[same format rules 1--4]}\\
\texttt{5. End the entire solution with the phrase: `The final answer is: [Letter]'}\\
\texttt{\ \ \ where [Letter] is the single uppercase character corresponding to your choice.}\\
\texttt{[rules 6--7: no extra text, no repeated output]}\\[4pt]
\texttt{Question: \{question\}}\\[4pt]
\texttt{Solution:}
\end{mdframed}

\subsection{Experimental Protocol Details}
\label{sec:protocol}

\paragraph{Dataset splits and sample sizes.}
We use the standard test split for GSM8K, MATH, and StrategyQA, and the validation split for CSQA, HotpotQA, and MuSiQue. For each dataset we randomly sample 800 examples per model, with the exception of StrategyQA where the full test set of 687 examples is used.

\paragraph{Decoding.}
All CoT trajectories are generated with sampling (\texttt{temperature=0.7}, \texttt{top\_p=0.9}, \texttt{max\_new\_tokens=512}). Up to five retries are attempted per sample if the output fails format validation (i.e., does not contain a parseable ``The final answer is:'' terminator).

\paragraph{Correctness evaluation.}
For math benchmarks (GSM8K, MATH), predicted and ground-truth answers are normalized by stripping punctuation and currency symbols and casting to float, then compared by exact match. For CSQA and StrategyQA, answers are normalized to the choice letter (A--E) and True/False respectively, then compared by exact match. For open-ended multi-hop benchmarks (HotpotQA, MuSiQue), string normalization is applied and a substring containment check is used as a fallback.

\subsection{Dataset Statistics and Class Balance}
\label{sec:class_balance}

Table~\ref{tab:class_balance} reports the number of evaluated samples and the fraction of
correct trajectories for each dataset--model combination. Correctness rates vary substantially
across datasets and models, from 5.0\% on MuSiQue (Phi-4-mini) to 83.9\% on GSM8K (Gemma-3-4B),
motivating the use of stratified splits and threshold tuning in all classification experiments.

\begin{table}[h]
\centering
\small
\caption{Sample sizes and correct trajectory rates per dataset and model.}
\label{tab:class_balance}
\begin{tabular}{llrr}
\toprule
\textbf{Dataset} & \textbf{Model} & \textbf{N} & \textbf{\% Correct} \\
\midrule
GSM8K       & Gemma-3-4B           & 800 & 83.9 \\
GSM8K       & LLaMA-3.2-3B         & 800 & 65.2 \\
GSM8K       & Phi-4-mini           & 800 & 55.1 \\
\midrule
MATH        & Gemma-3-4B           & 800 & 49.1 \\
MATH        & LLaMA-3.2-3B         & 800 & 34.1 \\
MATH        & Phi-4-mini           & 800 & 38.0 \\
\midrule
CSQA        & Gemma-3-4B           & 800 & 69.6 \\
CSQA        & LLaMA-3.2-3B         & 800 & 70.4 \\
CSQA        & Phi-4-mini           & 800 & 70.1 \\
\midrule
StrategyQA  & Gemma-3-4B           & 687 & 69.7 \\
StrategyQA  & LLaMA-3.2-3B         & 687 & 59.1 \\
StrategyQA  & Phi-4-mini           & 687 & 64.6 \\
\midrule
HotpotQA    & Gemma-3-4B           & 800 & 17.0 \\
HotpotQA    & LLaMA-3.2-3B         & 800 & 17.1 \\
HotpotQA    & Phi-4-mini           & 800 & 16.4 \\
\midrule
MuSiQue     & Gemma-3-4B           & 800 &  7.5 \\
MuSiQue     & LLaMA-3.2-3B         & 800 &  8.2 \\
MuSiQue     & Phi-4-mini           & 800 &  5.0 \\
\bottomrule
\end{tabular}
\end{table}

\begin{table*}[h!]
\centering
\small
\caption{F1 Comparison: SARE (w/o Token Count) vs. All Baselines for \textbf{LLaMA-3.2-3B}}
\label{tab:f1_notoken_llama_3_2_3b}
\begin{tabular*}{\linewidth}{@{\extracolsep{\fill}}lc|ccccc}
\toprule
Dataset & SARE & Token & LogProb & NegEnt & NegPerp & SelfCert \\
\midrule
GSM8K      & \textbf{0.527} & 0.514 & 0.512 & 0.507 & 0.512 & 0.505 \\
MATH       & \textbf{0.801} & 0.789 & 0.792 & 0.792 & 0.792 & 0.792 \\
CSQA       & \textbf{0.454} & 0.448 & 0.436 & 0.431 & 0.436 & 0.423 \\
HotpotQA   & \textbf{0.900} & 0.900 & \textbf{0.900} & \textbf{0.900} & \textbf{0.900} & \textbf{0.900} \\
MuSiQue    & \textbf{0.958} & 0.947 & 0.954 & 0.954 & 0.954 & 0.947 \\
StrategyQA & 0.536 & \textbf{0.580} & 0.000 & 0.000 & 0.000 & 0.000 \\
\bottomrule
\end{tabular*}
\end{table*}

\begin{table*}[h!]
\centering
\small
\caption{F1 Comparison: SARE (w/o Token Count) vs. All Baselines for \textbf{Gemma-3-4B}}
\label{tab:f1_notoken_gemma_3_4b}
\begin{tabular*}{\linewidth}{@{\extracolsep{\fill}}lc|ccccc}
\toprule
Dataset & SARE & Token & LogProb & NegEnt & NegPerp & SelfCert \\
\midrule
GSM8K      & \textbf{0.305} & 0.200 & 0.237 & 0.262 & 0.236 & 0.282 \\
MATH       & \textbf{0.717} & 0.691 & 0.640 & 0.634 & 0.640 & 0.673 \\
CSQA       & \textbf{0.473} & 0.471 & 0.451 & 0.415 & 0.451 & 0.465 \\
HotpotQA   & \textbf{0.893} & 0.885 & \textbf{0.893} & 0.864 & \textbf{0.893} & 0.889 \\
MuSiQue    & \textbf{0.964} & \textbf{0.964} & 0.961 & \textbf{0.964} & 0.961 & \textbf{0.964} \\
StrategyQA & \textbf{0.462} & 0.460 & 0.000 & 0.000 & 0.000 & 0.000 \\
\bottomrule
\end{tabular*}
\end{table*}

\begin{table*}[h!]
\centering
\small
\caption{F1 Comparison: SARE (w/o Token Count) vs. All Baselines for \textbf{Phi-4-mini}}
\label{tab:f1_notoken_phi_4_mini}
\begin{tabular*}{\linewidth}{@{\extracolsep{\fill}}lc|ccccc}
\toprule
Dataset & SARE & Token & LogProb & NegEnt & NegPerp & SelfCert \\
\midrule
GSM8K      & \textbf{0.628} & 0.617 & 0.615 & 0.609 & 0.615 & 0.615 \\
MATH       & \textbf{0.769} & \textbf{0.769} & 0.755 & 0.764 & 0.755 & 0.755 \\
CSQA       & 0.462 & \textbf{0.464} & 0.462 & 0.462 & 0.462 & \textbf{0.464} \\
HotpotQA   & \textbf{0.908} & 0.907 & 0.904 & \textbf{0.908} & 0.904 & 0.891 \\
MuSiQue    & \textbf{0.974} & 0.968 & \textbf{0.974} & \textbf{0.974} & \textbf{0.974} & 0.968 \\
StrategyQA & 0.494 & \textbf{0.514} & 0.000 & 0.000 & 0.000 & 0.000 \\
\bottomrule
\end{tabular*}
\end{table*}

\newpage
\begin{table*}[h!]
\centering
\small
\caption{AUPRC Comparison: SARE vs. All Baselines for \textbf{LLaMA-3.2-3B}}
\label{tab:auprc_full_llama_3_2_3b}
\begin{tabular*}{\linewidth}{@{\extracolsep{\fill}}lc|ccccc}
\toprule
Dataset & SARE & Token & LogProb & NegEnt & NegPerp & SelfCert \\
\midrule
GSM8K      & \textbf{0.479} & 0.420 & 0.286 & 0.257 & 0.286 & 0.246 \\
MATH       & \textbf{0.760} & 0.728 & 0.585 & 0.560 & 0.585 & 0.548 \\
CSQA       & 0.329 & \textbf{0.331} & 0.242 & 0.251 & 0.242 & 0.252 \\
HotpotQA   & 0.861 & \textbf{0.869} & 0.730 & 0.708 & 0.730 & 0.698 \\
MuSiQue    & \textbf{0.921} & 0.900 & 0.881 & 0.870 & 0.881 & 0.868 \\
StrategyQA & 0.382 & \textbf{0.451} & 0.000 & 0.000 & 0.000 & 0.000 \\
\bottomrule
\end{tabular*}
\end{table*}

\begin{table*}[h!]
\centering
\small
\caption{AUPRC Comparison: SARE vs. All Baselines for \textbf{Gemma-3-4B}}
\label{tab:auprc_full_gemma_3_4b}
\begin{tabular*}{\linewidth}{@{\extracolsep{\fill}}lc|ccccc}
\toprule
Dataset & SARE & Token & LogProb & NegEnt & NegPerp & SelfCert \\
\midrule
GSM8K      & \textbf{0.217} & 0.199 & 0.159 & 0.156 & 0.159 & 0.210 \\
MATH       & \textbf{0.838} & 0.779 & 0.511 & 0.539 & 0.511 & 0.600 \\
CSQA       & \textbf{0.500} & 0.454 & 0.318 & 0.280 & 0.318 & 0.374 \\
HotpotQA   & \textbf{0.837} & 0.813 & 0.718 & 0.731 & 0.718 & 0.751 \\
MuSiQue    & \textbf{0.926} & 0.909 & 0.893 & 0.917 & 0.893 & 0.904 \\
StrategyQA & \textbf{0.419} & 0.390 & 0.000 & 0.000 & 0.000 & 0.000 \\
\bottomrule
\end{tabular*}
\end{table*}

\begin{table*}[h!]
\centering
\small
\caption{AUPRC Comparison: SARE vs. All Baselines for \textbf{Phi-4-mini}}
\label{tab:auprc_full_phi_4_mini}
\begin{tabular*}{\linewidth}{@{\extracolsep{\fill}}lc|ccccc}
\toprule
Dataset & SARE & Token & LogProb & NegEnt & NegPerp & SelfCert \\
\midrule
GSM8K      & \textbf{0.644} & 0.395 & 0.380 & 0.333 & 0.380 & 0.312 \\
MATH       & \textbf{0.702} & 0.600 & 0.528 & 0.504 & 0.528 & 0.498 \\
CSQA       & 0.372 & \textbf{0.428} & 0.260 & 0.246 & 0.260 & 0.225 \\
HotpotQA   & \textbf{0.894} & 0.882 & 0.770 & 0.755 & 0.770 & 0.756 \\
MuSiQue    & \textbf{0.969} & 0.958 & 0.964 & 0.946 & 0.964 & 0.940 \\
StrategyQA & \textbf{0.367} & 0.347 & 0.000 & 0.000 & 0.000 & 0.000 \\
\bottomrule
\end{tabular*}
\end{table*}
\label{sec:appendix}


\end{document}